\documentclass[11pt,letterpaper]{article}
\usepackage{emnlp2016}
\usepackage{times}
\usepackage{latexsym}
\usepackage{subfigure}%,subcaption}
\usepackage{amsmath,graphicx} 
\usepackage{amssymb}
\usepackage{multirow}
\usepackage{hhline}

\emnlpfinalcopy

\def\emnlppaperid{364}

\title{Deep Reinforcement Learning with a Combinatorial Action Space for Predicting Popular Reddit Threads}

\author{%
Ji He$^*$, Mari Ostendorf$^*$, Xiaodong He$^\dagger$, Jianshu Chen$^\dagger$, Jianfeng Gao$^\dagger$, Lihong Li$^\dagger$, Li Deng$^\dagger$ \\
$^*$Department of Electrical Engineering,
University of Washington,
Seattle, WA 98195, USA \\
\texttt{\{jvking, ostendor\}@uw.edu}
\\
$^\dagger$Microsoft Research,
Redmond, WA 98052, USA \\
\texttt{\{xiaohe, jianshuc, jfgao, lihongli, deng\}@microsoft.com}
}

\begin{document}

\maketitle

\begin{abstract}
We introduce an online popularity prediction and tracking task as a benchmark task for reinforcement learning with a combinatorial, natural language action space. A specified number of discussion threads predicted to be popular are recommended, chosen from a fixed window of recent comments to track. Novel deep reinforcement learning architectures are studied for effective modeling of the value function associated with actions comprised of \emph{interdependent} sub-actions. The proposed model, which represents dependence between sub-actions through a bi-directional LSTM, gives the best performance across different experimental configurations and domains, and it also generalizes well with varying numbers of recommendation requests.
%We propose a Reddit popularity prediction and tracking task as a benchmark task for reinforcement learning with a natural language action space. Popular discussion threads are predicted and recommended, and recommending multiple new comments is a combinatorial action. Novel architectures are studied for better modeling value function associated with actions combined with interdependent sub-actions. The proposed DRRN-BiLSTM can not only perform better across different experimental configurations and various domains, but also generalize well when user request changes.
%(need a total rewrite) This paper presents a novel architecture for handling a combinatorial action space in reinforcement learning setting, with states and actions represented in natural language. In situations where a complex action is a combination of multiple sub-actions, we use a bi-directional Long Short-Term Memory (LSTM) to model the action Q-value functions. The exponentially complexity of a combinatorial action space also poses challenges such as enumerating over all possible action choices. The proposed DRRN-BiLstm is able to capture the structure in order to learn an efficient function approximation. Experiments with Reddit popularity tracking task show that the model outperforms significantly over baseline Per-action DQN and simple addition of DRRN Q-values.
\end{abstract}

\section{Introduction}
\label{sec:intro}

This paper is concerned with learning policies for sequential decision-making tasks, where a system takes actions given options characterized by natural language with the goal of maximizing a long-term reward. More specifically, we consider tasks with a combinatorial action space, where each action is a set of multiple interdependent sub-actions.
%Furthermore, the states and actions are characterized with natural language text, making both state and action space large and discrete. 
The problem of a combinatorial natural language action space arises in many applications. For example, in real-time news feed recommendation, a user may want to read diverse topics of interest, and an action (i.e. recommendation) from the computer agent would consist of a set of news articles that are not all similar in topics \cite{yue2011linear}. In advertisement placement, an action is a selection of several ads to display, and bundling with complementary products might receive higher click-through-rate than displaying all similar popular products.
% In resource allocation, assigning multiple workloads to employees also involves cost computation over a combined choice.

% In webpage navigation \cite{nogueira2016webnav}, an agent keeps track of more than one path in order to search for a matched query, and the action is to select multiple hyperlinks among all the currently presented candidate hyperlinks. 
%%MO: took this out based on Jianfeng's comments, added alternative applications below in trying to distinguish our application, but getting another example here to replace CRM would still be good
%In customer relationship management, the action could be a combination of sub-actions, such as sending a particular follow-up email, assigning a technical solution professional, or offering one of several possible discounts.

In this work, we consider Reddit popularity prediction, which is similar to newsfeed recommendation but different in two respects. First, our goal is not to make recommendations based on an individual's preferences, but instead based on the anticipated long-term interest level of a broad group of readers from a target community. Second, we try to predict rather than detect popularity. Unlike individual interests, community interest level is not often immediately clear; there is a time lag before the level of interest starts to take off. Here, the goal is for the recommendation system to identify and track written documents (e.g. news articles, comments in discussion forum threads, or scientific articles) in real time -- attempting to identify hot updates before they become hot to keep the reader at the leading edge. The premise is that the user's bandwidth is limited, and only a limited number of things can be recommended out of several possibilities. In our experimental work, we use discussion forum text, where the recommendations correspond to recent posts or comments, assessing interest based on community response as observed in ``likes'' or other positive reactions to those comments. For training purposes, we can use community response measured at a time much later than the original post or publication. This problem is well-suited to the reinforcement learning paradigm, since the reward (the level of community uptake or positive response) is not immediately known, so the system needs to learn a mechanism for estimating future reactions. Different from typical reinforcement learning, the action space is combinatorial since an action corresponds to a set of comments (sub-actions) chosen from a larger set of candidates. A sub-action is a written comment (or document, for another variant of this task). 
%%MOnew: I make this point in section 2 instead
%An advantage of working with discussion forum comments is that the comments represent relatively small documents which reduces the scale of the problem for initial development.

Two challenges associated with this problem include the potentially high computational complexity of the combinatorial action space and the development of a framework for estimating the long-term reward (the Q-value in reinforcement learning) from a combination of sub-actions characterized by natural language. Here, we focus on the second problem, exploring different deep neural network architectures in an effort to efficiently account for the potential redundancy and/or temporal dependency of different sub-actions in relation to the state space. We sidestep the computational complexity issue (for now) by working with a task where the number of combinations is not too large and by further reducing costs by random sampling.  

There are two main contributions in this paper. First, we propose a novel reinforcement learning task with both states and combinatorial actions defined by natural language,\footnote{Simulator code and Reddit discussion identifiers are released at \tt{https://github.com/jvking/reddit-RL-
simulator}} which is introduced in section~\ref{sec:task}. This task, which is  based on comment popularity prediction using data from the Reddit discussion forum, can serve as a benchmark in social media recommendation and trend spotting.    The second contribution is the development of a novel deep reinforcement learning architecture for handling a combinatorial action space associated with natural language. Prior work related to both the task and deep reinforcement learning is reviewed in section~\ref{sec:related}, 
%The approach builds on prior work in deep reinforcement learning summarized in section~\ref{sec:related}. 
Details for the new models and baseline architectures are described in section~\ref{sec:model}. Experimental results in section~\ref{sec:experiments} show the proposed methods 
%capture the combinatorial structure and outperform baseline models, with DRRN-BiLSTM performing consistently the best. 
outperform baseline models and that a bidirectional LSTM is effective for characterizing the combined utility of sub-actions.
A brief summary of findings and open questions are in section~\ref{sec:concl}.

%%MO: This needs to go into related work not intro -- you haven't introduced enough concepts yet, fix the ``understanding'' claim when it is moved
%Different from many text-based games \cite{narasimhan-kulkarni-barzilay:2015:EMNLP,he2016deep} and a webpage navigation task \cite{nogueira2016webnav}, we explicitly separate train/test data. Thus during the testing stage, the agent will not receive the same states or actions that appear in the online training stage. An agent needs to gain a decent level of language understanding in order to perform well in both online training and testing.

\section{Popularity Prediction and Tracking}
\label{sec:task}
Our experiments are based on Reddit\footnote{\tt{http://www.reddit.com}}, one of the world's largest public discussion forums. On Reddit, registered users initiate a post and people respond with comments, either to the original post or one of its associated comments.  Together, the comments and the original post form a discussion tree, which grows as new comments are contributed.
%As time goes on, more comments will appear and the tree continues to grow. 
%%MOnew: moved weninger here since it is not about topic tracking
It has been show that discussions tend to have a hierarchical topic structure \cite{weninger2013exploration}, i.e.\ different branches of the discussion reflect narrowing of higher level topics. 
Reddit discussions are grouped into different domains, called subreddits, according to different topics or themes. Depending on the popularity of the subreddit, a post can receive hundreds of comments.  

\begin{figure}[t]
\centerline{
		\includegraphics[width=0.41\textwidth]{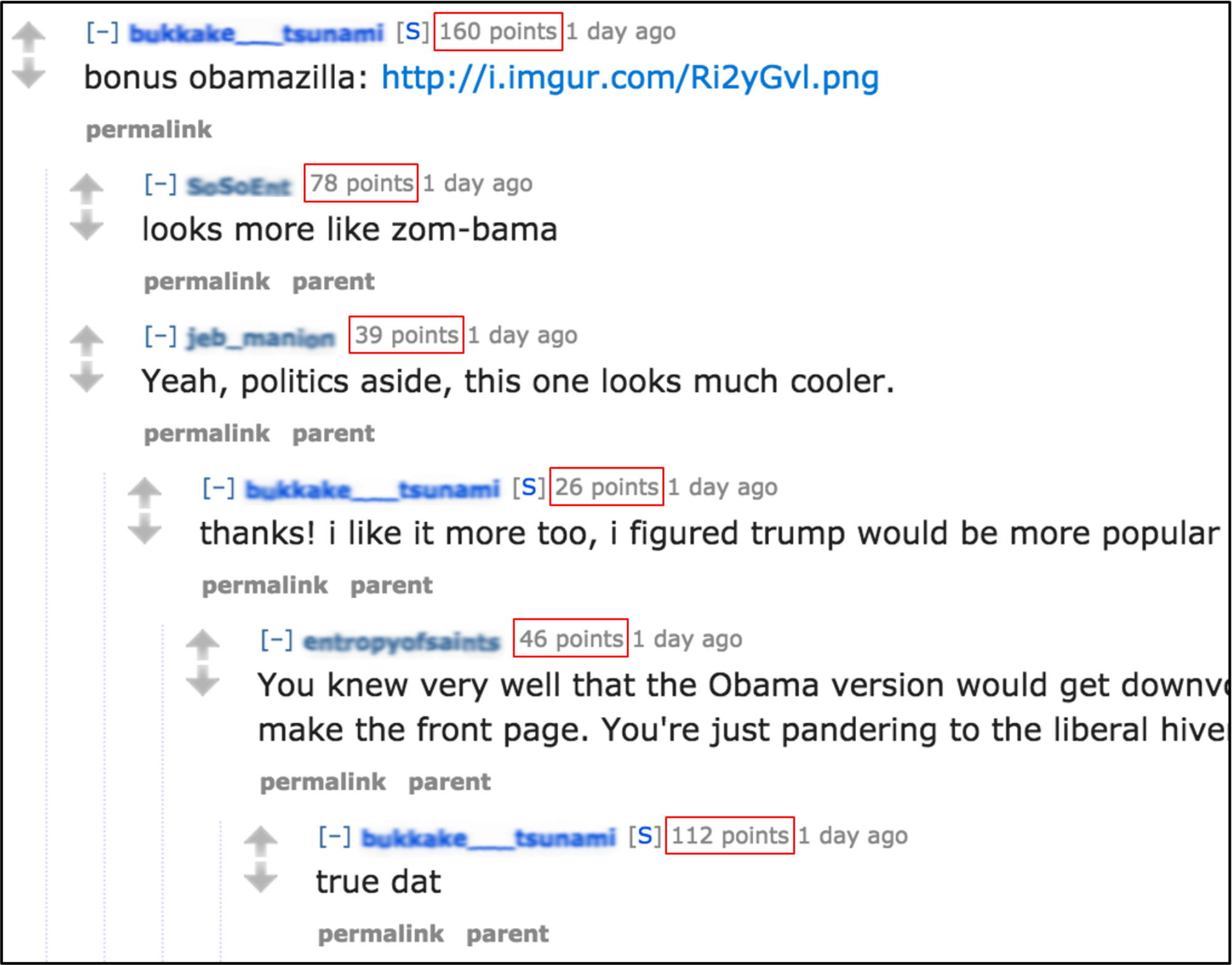}
 } 
  \caption{A snapshot of the top of a Reddit discussion tree, where karma scores are shown in red boxes.}
  \label{Fig:Reddit-example}
\end{figure}

Comments (and posts) are associated with positive and negative votes (i.e., likes and dislikes) from registered users that are combined to get a \emph{karma score}, which can be used as a measure for popularity. An example of the top of a Reddit discussion tree is given in Figure \ref{Fig:Reddit-example}. The scores in red boxes mark the current karma (popularity) of each comment, and it is quite common that a lower karma comment (e.g. ``Yeah, politics aside, this one looks much cooler'', compared to ``looks more like zom-bama'') will lead to more children and popular comments in the future (e.g. ``true dat''). Note that the karma scores are dynamic, changing as readers react to the evolving discussion and eventually settling down as the discussion trails off. In a real-time comment recommendation system, the eventual karma of a comment is not immediately available, so prediction of popularity is based on the text in the comment in the context of prior comments in the subtree and other comments in the current time window.

Popularity prediction and tracking in the Reddit setting is used in this paper for studying reinforcement learning to model long-term rewards in a combinatorial action space.
At each time step, the state corresponds to the collection of comments previously recommended. The system aims at automatically picking a few lines of the discussion to follow from the new set of comments in a given window, which is a combinatorial action. 
%%MOnew:
Thread popularity tracking can be thought of as a proxy task for news or scientific article recommendation.  It has the advantages that ``documents'' (comments) are relatively short and that the long-term reward can be characterized by Reddit voting scores, which makes this task easier to work with for algorithm development than these larger related tasks.

In this work, we only consider new comments associated with the threads of the discussion that we are currently following to limit the number of possible sub-actions at each time step and with the assumption that prior context is needed to interpret the comments. In other words, the new recommendation should focus on comments that are in the subtrees of previously recommended comments.  (A variant relaxing this restriction is suggested in the conclusion section.)  
Typically, one would expect some interdependencies between comments made in the same window if they fall under the same subtree, because they correspond to a reply to the same parent. In addition, there may be some temporal dependency, since one sub-action may be a comment on the other. These dependencies will affect the combined utility of the sub-actions.

According to our experiments, the performance is significantly worse when we learn a myopic policy compared to reinforcement learning with the same feature set. This shows that long-term dependency indeed matters, as illustrated in Figure \ref{Fig:Reddit-example}. This serves as a justification that reinforcement learning is an appropriate approach for modeling popularity of a discussion thread.

%According to our experiments, greedily picking direct replies with highest karma only gives 50\% to 60\% of the upperbound\footnote{Upperbounds are estimated by greedily searching through each discussion tree to find $K$ max Karma discussion threads (overlapped comments are counted only once). This upper bound may not be attainable in real-time setting.} if we do a long term search in an offline constructed tree. The performance is also significantly worse when we tried learning a myopic policy compared to reinforcement learning using the same feature set. This shows that long-term dependency indeed matters and an example is given in Figure \ref{Fig:Reddit-example}. This serves as a justification that reinforcement learning is an appropriate approach for modeling popularity of a discussion thread.

\section{Related Work}
\label{sec:related}

There is a large body of work on reinforcement learning. Among those of most interest here are deep reinforcement learning methods that leverage neural networks because of their success in handling large discrete state/action spaces. Early work such as TD-gammon used a neural network to approximate the state value function \cite{tesauro1995temporal}. Recent advances in deep learning \cite{lecun2015deep,deng2014deep,hinton2012deep,krizhevsky2012imagenet,sordoni2015neural} inspired significant progress by combining deep learning with reinforcement learning \cite{mnih2015human,silver2016mastering,lillicrap2015continuous,duan2016benchmarking}. In natural language processing, reinforcement learning has been applied successfully to dialogue systems that generate natural language and converse with a human user \cite{scheffler2002automatic,singh1999reinforcement,wen2016network}. There has also been interest in mapping text instructions to sequences of executable actions and extracting textual knowledge to improve game control performance \cite{branavan-EtAl:2009:ACLIJCNLP,branavan2011learning}.

Recently, Narasimhan et al.\ \shortcite{narasimhan-kulkarni-barzilay:2015:EMNLP} studied the task of text-based games with a deep Q-learning framework. He et al. \shortcite{he2016deep} proposed to use a separate deep network for handling natural language actions and to model Q-values via state-action interaction. Nogueira and Cho \shortcite{nogueira2016webnav} have also proposed a goal-driven web navigation task for language-based sequential decision making. Narasimhan et al. \shortcite{narasimhan2016improving} applied reinforcement learning for acquiring and incorporating external evidence to improve information extraction accuracy. The study that we present with Reddit popularity tracking differs from these other text-based reinforcement learning tasks in that the language in both state and action spaces is unconstrained and quite rich.

%%MOnew: some revisions for readability in this para
Dulac-Arnold et al.\ \shortcite{dulacdeep} also investigated a problem of large discrete action spaces. A Wolpertinger architecture is proposed to reduce computational complexity of evaluating all actions. While a combinatorial action space can be large and discrete, their method does not directly apply in our case, because the possible actions are changing over different states. In addition, our work differs in that its focus is on modeling the combined action-value function rather than on reducing computational complexity. Other work that targets a structured action space includes: an actor-critic algorithm, where actions can have real-valued parameters \cite{hausknecht2015deep}; and the factored Markov Decision Process (MDP) \cite{guestrin2001multiagent,sallans2004reinforcement}, with certain independence assumptions between a next-state component and a sub-action. As for a bandits setting, Yue and Guestrin \shortcite{yue2011linear} considered diversification of multi-item recommendation, but their methodology is limited to using linear approximation with hand-crafted features.

%%MO: the Weninger paper is not about topic tracking. It's sbout showing that threaded discussions tend to exhibit a topical hierarchy. I put the ref into the task section

%%MOnew;
The task explored in our paper -- detecting and tracking popular threads in a discussion -- is somewhat related to topic detection and tracking \cite{allan2012topic,mathioudakis2010twittermonitor}, but it differs in that the goal is not to track topics based on frequency, but rather based on reader response. Thus, our work is more closely related to popularity prediction for social media and online news. These studies have explored a variety of definitions (or measurements) of popularity, including: the volume of comments in response to blog posts \cite{Yano2010ICWSM} and news articles \cite{Tsagkias+09,Tatar+11}, the number of Twitter shares of news articles \cite{Bandari+12}, the number of reshares on Facebook \cite{Cheng+14} and retweets on Twitter \cite{Suh+10,Hong+11,Tan2014ACL,Zhao2015KDD}, the rate of posts related to a source rumor \cite{Lukasik+15}, and the difference in the number of reader up and down votes on posts and comments in Reddit discussion forums \cite{Lakkaraju2013ICWSM,Jaech2015}. An advantage of working with the Reddit data is that both positive and negative reactions are accounted for in the karma score. Of the prior work on Reddit, the task explored here is most similar to \cite{Jaech2015} in that it involves choosing relatively high karma comments (or threads) from a time-limited set rather than directly predicting comment (or post) karma. Prior work on popularity prediction used supervised learning; this is the first work that frames tracking hot topics in social media with deep reinforcement learning.

\section{Characterizing a combinatorial action space}
\label{sec:model}

\subsection{Notation}
In this sequential decision making problem, at each time step $t$, the agent receives a text string that describes the state $s_t\in\mathcal{S}$ (i.e., ``state-text'') and picks a text string that describes the action $a_t\in\mathcal{A}$ (i.e., ``action-text''), where $\mathcal{S}$ and $\mathcal{A}$ denote the state and action spaces, respectively. Here, we assume $a_t$ is chosen from a set of given candidates. In our case both $\mathcal{S}$ and $\mathcal{A}$ are described by natural language. Given the state-text and action-texts, the agent aims to select the best action in order to maximize its long-term reward. Then the environment state is updated $s_{t+1}=s'$ according to a probability $p(s'|s,a)$, and the agent receives a reward $r_{t+1}$ for that particular transition. We define the action-value function (i.e.\ the Q-function) $Q(s,a)$ as the expected return starting from $s$ and taking the action $a$:
\begin{align*}
Q(s,a)=\mathbb{E}\left\{\sum_{l=0}^{+\infty}\gamma^l r_{t+1+l} | s_t=s, a_t=a \right\}
\end{align*}
where $\gamma\in (0,1)$ denotes a discount factor. The Q-function associated with an optimal policy can be found by the Q-learning algorithm \cite{watkins1992q}:
\begin{align*}
	Q(s_t, a_t) 
		&\leftarrow
			Q(s_t, a_t) 
			+ \\ \nonumber
		&	\eta_t \cdot 
			\big(
				r_{t+1} + \gamma \cdot \max_{a}{Q(s_{t+1}, a)}-Q(s_t, a_t)
			\big) 
\end{align*}
where $\eta_t$ is a learning rate parameter.

The set of comments that are being tracked at time step $t$ is denoted as $M_t$. All previously tracked comments, as well as the post (root node of the tree), is considered as state $s_t$ ($s_t=\{M_0, M_1, \cdots, M_t\}$), and we initialize $s_0=M_0$ to be the post. An action is taken when a total of $N$ new comments $\{c_{t,1}, c_{t,2}, \cdots, c_{t,N}\}$ appear as nodes in the subtree of $M_t$, and the agent picks a set of $K$ comments to be tracked in the next time step $t+1$. Thus we have:
\begin{align}
a_t=\{c_{t}^{1}, c_{t}^{2}, \cdots, c_{t}^{K}\},\quad& c_{t}^{i} \in \{c_{t,1}, c_{t,2}, \cdots, c_{t,N}\} \nonumber \\
&\text{and } c_{t}^{i} \ne c_{t}^{j} \text{ if } i\ne j \label{Equ:combine-sub-action}
\end{align}
and $M_{t+1}=a_t$. At the same time, by taking action $a_t$ at state $s_t$, the reward $r_{t+1}$ is the accumulated karma scores, i.e. sum over all comments in $M_{t+1}$. Note that the reward signal is used in online training, while at model deployment (testing stage), the scores are only used as an evaluation metric.

Following the reinforcement learning tradition, we call tracking of a single discussion tree from start (root node post) to end (no more new comments appear) an \emph{episode}. We also randomly partition all discussion trees into separate training and testing sets, so that texts seen by the agent in training and testing are from the same domain but different discussions. For each episode, depending on whether training/testing, the simulator randomly picks a discussion tree, and presents the agent with the current state and $N$ new comments.

%\begin{figure*}[t]\centering
%  	\begin{subfigure}[b]{0.45\textwidth}
%		\centering
%		\includegraphics[width=0.74\textwidth]{architecture_pa-dqn_v3.pdf}
%		\caption{Per-action DQN}
%		\label{fig:architectures_pa-dqn}
%	\end{subfigure}
%%  	\hfill
%  	\begin{subfigure}[b]{0.45\textwidth}
%		\centering
%		\includegraphics[width=0.8\textwidth]{architecture_DRRN_v3.pdf}
%		\caption{DRRN}
%		\label{fig:architectures_drrn}
%	\end{subfigure}
%	\vskip\baselineskip
%  	\begin{subfigure}[b]{0.45\textwidth}
%		\centering
%		\includegraphics[width=0.95\textwidth]{architecture_DRRN-Sum_v3.pdf}
%		\caption{DRRN-Sum}
%		\label{fig:architectures_drrn-sum}
%	\end{subfigure}
%%  	\hfill
%  	\begin{subfigure}[b]{0.45\textwidth}
%		\centering
%		\includegraphics[width=0.81\textwidth]{architecture_DRRN-BiLstm_v4.pdf}
%		\caption{DRRN-BiLSTM}
%		\label{fig:architectures_drrn-bilstm}
%	\end{subfigure}
%\caption{Different deep Q-learning architectures}
%\label{fig:architectures}
%\end{figure*}

\begin{figure*}[t]
  \centerline{
  	\hfill
	\subfigure[Per-action DQN]{
	\includegraphics[width=0.35\textwidth]{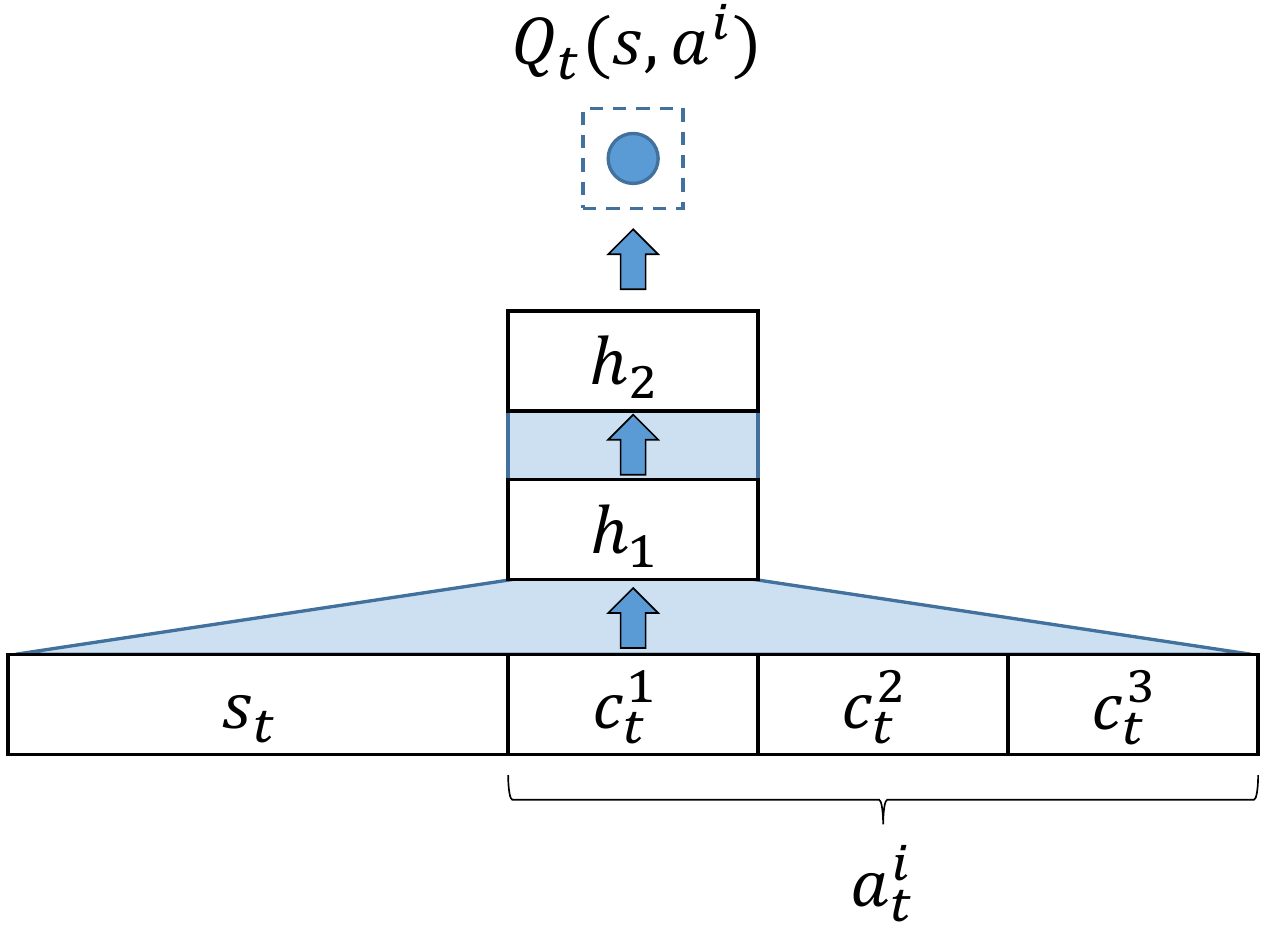}
	\label{fig:architectures_pa-dqn}
  	}
  	\hfill
	\hfill
	\subfigure[DRRN]{
	\includegraphics[width=0.38\textwidth]{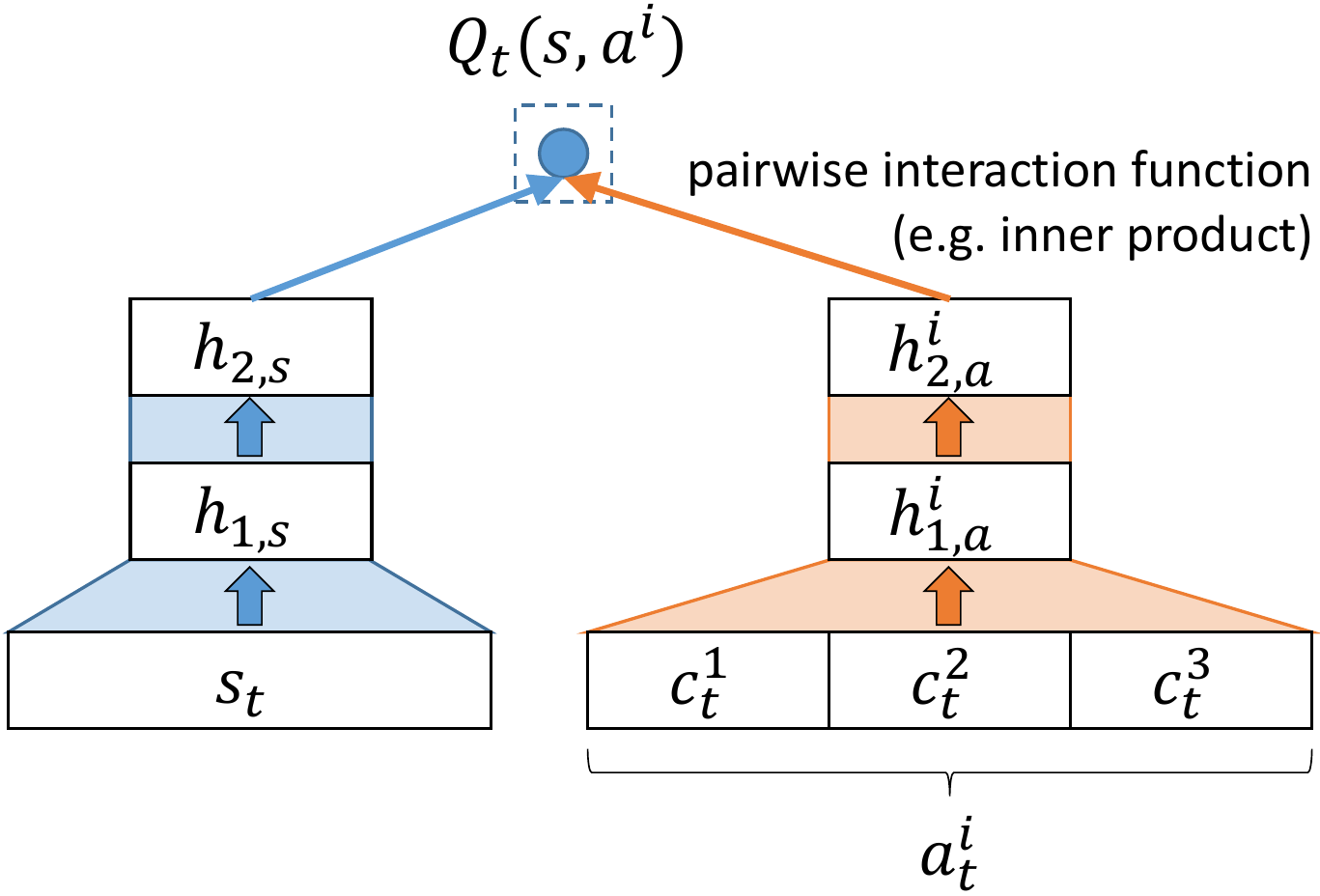}
  	\label{fig:architectures_drrn}
  	}
	\hfill
	}
  \centerline{
  	\hfill
	\subfigure[DRRN-Sum]{
	\includegraphics[width=0.46\textwidth]{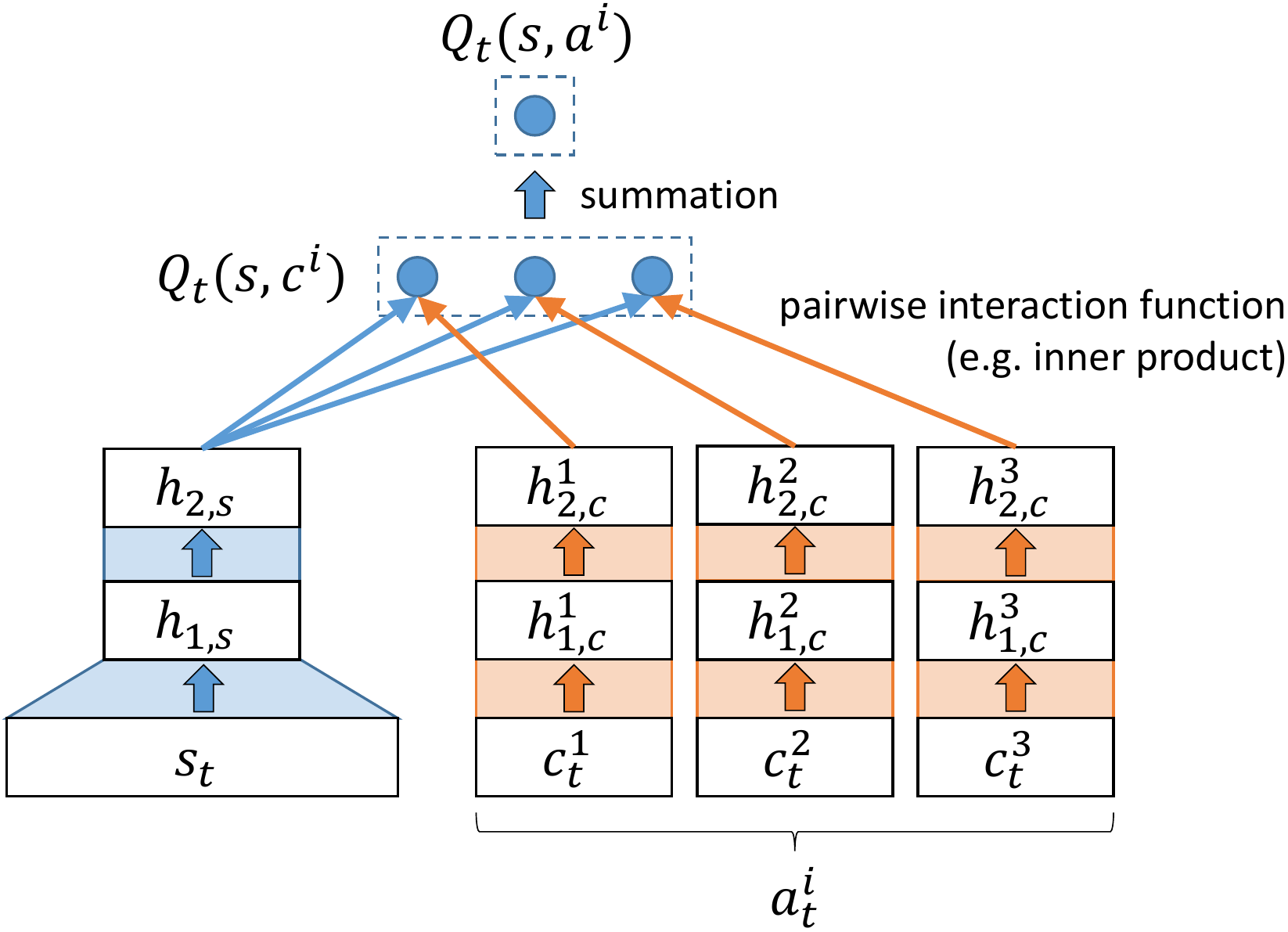}
  	\label{fig:architectures_drrn-sum}
  	}
  	\hfill
	\hfill
	\subfigure[DRRN-BiLSTM]{
	\includegraphics[width=0.39\textwidth]{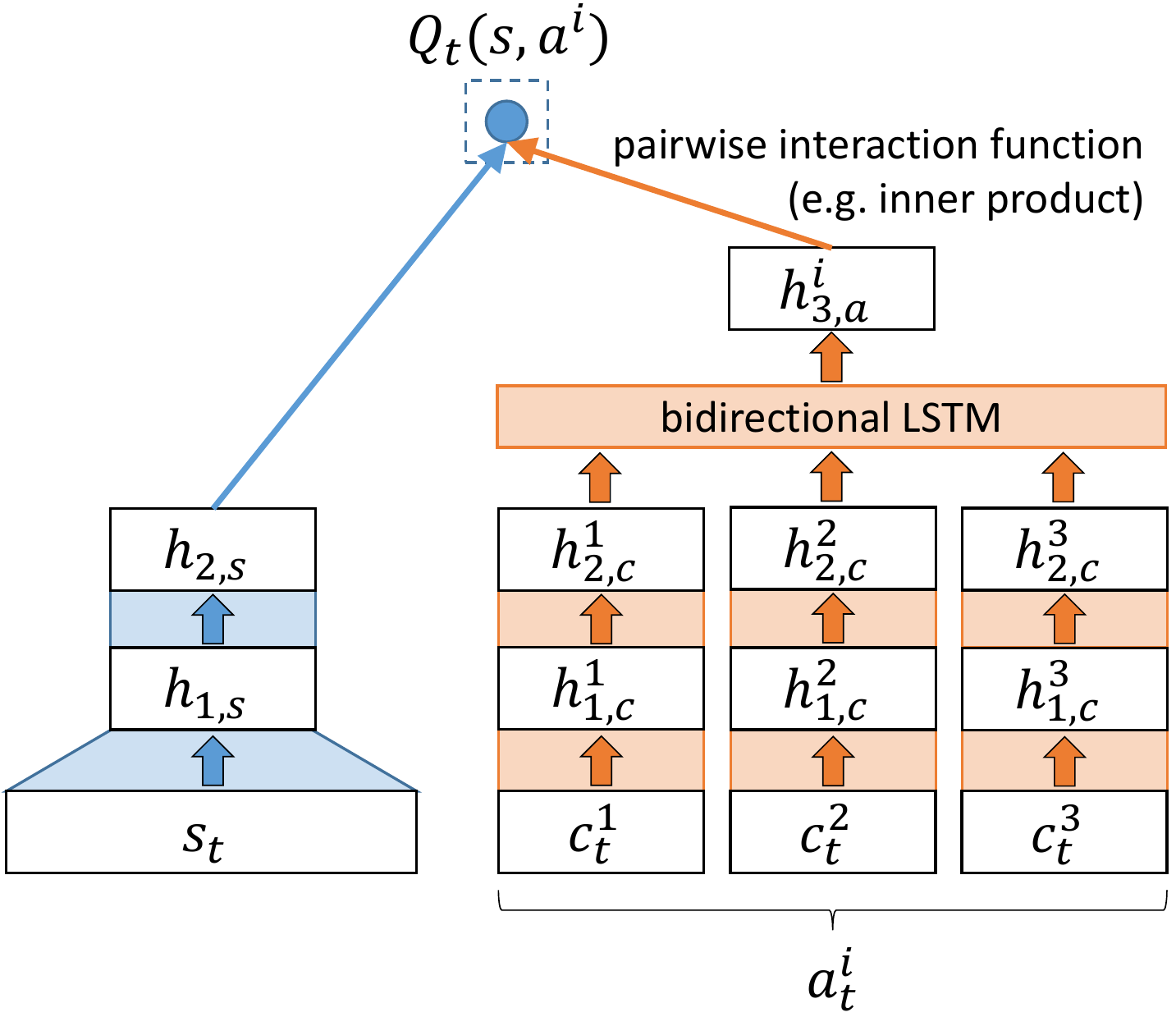}
	\label{fig:architectures_drrn-bilstm}
  	}
	\hfill
	\hfill
  }
\caption{Different deep Q-learning architectures}
\label{fig:architectures}
\end{figure*}

%\subsection{Reddit popularity simulator}
%
%%%MO: I haven't finished cleaning this up
%
%We build a simulator that mimics the real-time popularity tracking scenario given a pre-filtered Reddit database. 
%
%At each time step $t$ in this episode, the simulator stores a set of comments that are being tracked, denoted as $M_t$. All previously tracked comments, as well as the post (root node of the tree), is considered as state $s_t$. Thus $s_t=\{M_0, M_1, \cdots, M_t\}$ and we initialize $s_0=M_0$ to be the post. The simulator waits until a total of $N$ new comments $\{c_{t,1}, c_{t,2}, \cdots, c_{t,N}\}$ appear as nodes in the subtree of $M_t$, and the action by the agent at time $t$ is to pick a set of $K$ comments to be tracked in the next time step $t+1$.  At the same time, by taking action $a_t$ at state $s_t$, the simulator presents accumulated karma scores as reward $r_{t+1}$ for this state transition. 

\subsection{Q-function alternatives}

%%MO: this was from the ``architectures'' section which was not intended to be a section. Make sure this material is merged in appropriately. I added figure pointers for DQN and DRRN
%

With the real-time setting, it is clear that action $a_t$ will affect the next state $s_{t+1}$ and furthermore the future expected reward. The action $a_t$ consists of $K$ comments (sub-actions), making modeling Q-values $Q(s_t,a_t)$ difficult. To handle a large state space, Mnih et al. \shortcite{mnih2015human} proposed a Deep Q-Network (DQN). In case of a large action space, we may use both state and action representations as input to a deep neural network. It is shown that the Deep Reinforcement Relevance Network (DRRN, Figure \ref{fig:architectures_drrn}), i.e. two separate deep neural networks for modeling state embedding and action embedding, performs better than per-action DQN (PA-DQN in Figure \ref{fig:architectures_pa-dqn}), as well as other DQN variants for dealing with natural language action spaces \cite{he2016deep}.

Our baseline models include Linear, PA-DQN and DRRN. We concatenate the $K$ sub-actions/comments to form the action representation. The Linear and PA-DQN (Figure \ref{fig:architectures_pa-dqn}) take as input a concatenation of state and action representations, and model a single Q-value $Q(s_t, a_t)$ using linear or DNN function approximations. The DRRN consists of a pair of DNNs, one for the state-text embedding and the other for action-text embeddings, which are then used to compute $Q(s_t, a_t)$ via a pairwise interaction function (Figure \ref{fig:architectures_drrn}).

One simple alternative approach by utilizing this combinatorial structure is to compute an embedding for each sub-action $c_t^i$. We can then model the value in picking a particular sub-action, $Q(s_t, c_t^i)$, through a pairwise interaction between the state and this sub-action. $Q(s_t, c_t^i)$ represents the expected accumulated future rewards by including this sub-action. The agent then greedily picks the top-$K$ sub-actions with highest values to achieve the highest $Q(s_t, a_t)$. In this approach, we are assuming the long-term rewards associated with sub-actions are independent of each other. More specifically, greedily picking the top-$K$ sub-actions is equivalent to maximizing the following action-value function:
\begin{align}
Q(s_t, a_t)=\sum_{i=1}^K Q(s_t, c_t^i)
\label{Equ:DRRN-Sum}
\end{align}
while satisfying \eqref{Equ:combine-sub-action}. We call this proposed method DRRN-Sum, and its architecture is shown in Figure \ref{fig:architectures_drrn-sum}. Similarly as in DRRN, we use two networks to embed state and actions separately. However, for different sub-actions, we keep the network parameters tied. We also use the same top layer dimension and the same pairwise interaction function for all sub-actions.

In the case of a linear additive interaction, such as an inner product or bilinear operation, Equation \eqref{Equ:DRRN-Sum} is equivalent to computing the interaction between the state embedding and an action embedding, where the action embedding is obtained linearly by summing over $K$ sub-action embeddings. When sub-actions have strong correlation, this independence assumption is invalid and can result in a poor estimation of $Q(s_t, a_t)$. For example, most people are interested in the total information stored in the combined action $a_t$. Due to content redundancy in the sub-actions $c_{t}^{1}, c_{t}^{2}, \cdots, c_{t}^{K}$, we expect $Q(s_t, a_t)$ to be smaller than $\sum_i Q(s_t, c_t^i)$.

To come up with a general model for handling a combinatorial action-value function, we further propose the DRRN-BiLSTM (Figure \ref{fig:architectures_drrn-bilstm}). In this architecture, we use a DNN to generate an embedding for each comment. Then a Bidirectional Long Short-Term Memory \cite{graves2005framewise} is used to combine a sequence of $K$ comment embeddings. As the Bidirectional LSTM has a larger capacity due to its nonlinear structure, we expect it will capture more details on how the embeddings for the sub-actions combine into an action embedding. Note that both of our proposed methods (DRRN-Sum and DRRN-BiLSTM) can handle a varying value of $K$, while for the DQN and DRRN baselines, we need to use a fixed $K$ in training and testing.

%The simple additive assumption in DRRN-Sum may work well in Reddit case, but does not apply to more sophisticated combinatorial action space. For example, sometimes people may be interested in the total information stored in this combined action $a_t$ while information gain is often a submodular set function. To propose a general architecture for handling a combinatorial action-value function, we also experiment with DRRN-Pooling (Figure \ref{fig:architectures_drrn-pooling}) and DRRN-BiLSTM (Figure \ref{fig:architectures_drrn-bilstm}). For both methods, we use a deep neural network

\section{Experiments}
\label{sec:experiments}

\subsection{Datasets and Experimental Configurations}
Our data consists of 5 subreddits (askscience, askmen, todayilearned, worldnews, nfl) with diverse topics and genres. In our experiments, in order to have long enough discussion threads, we filter out discussion trees with fewer than 100 comments. For each subreddit, we randomly partition 90\% of the data for online training, and 10\% of the data for testing (deployment). The basic subreddit statistics are shown in Table \ref{table:subreddit-stats}. We report the random policy performances and heuristic upper bound performances (averaged over 10,000 episodes) in Table \ref{table:subreddit-basic-performance} and Table \ref{table:askscience-basic-performance}.\footnote{Upper bounds are estimated by greedily searching through each discussion tree to find $K$ max karma discussion threads (overlapped comments are counted only once). This upper bound may not be attainable in a real-time setting.} The upper bound performances are obtained using stabilized karma scores and offline constructed tree structure. The mean and standard deviation are obtained by 5 independent runs.

\begin{table}
\small
\centering
\begin{tabular}{| l | c | c |} \hline
\bf Subreddit & \bf \# Posts (in k) & \bf \# Comments (in M) \\ \hline
askscience & 0.94 & 0.32 \\ \hline
askmen & 4.45 & 1.06 \\ \hline
todayilearned & 9.44 & 5.11 \\ \hline
worldnews & 9.88 & 5.99 \\ \hline
nfl & 11.73 & 6.12 \\ \hline
%\bf Total & 36.44 & 18.60 \\ \hline
\end{tabular}
\caption{Basic statistics of filtered subreddit data sets}
\label{table:subreddit-stats}
\end{table}

\begin{table}
\small
\centering
\begin{tabular}{| l | c | c |} \hline
\bf Subreddit & \bf Random & \bf Upper bound \\ \hline
%askscience & 321 (7) & 2109 (16) \\ \hline
%askmen & 132 (1) & 651 (3) \\ \hline
%todayilearned & 390 (6) & 2679 (30) \\ \hline
%worldnews & 206 (4) & 1853 (44) \\ \hline
%nfl & 237 (1) & 1338 (13) \\ \hline
askscience & 321.3 (7.0) & 2109.0 (16.5) \\ \hline
askmen & 132.4 (0.7) & 651.4 (2.8) \\ \hline
todayilearned & 390.3 (5.7) & 2679.6 (30.1) \\ \hline
worldnews & 205.8 (4.5) & 1853.4 (44.4) \\ \hline
nfl & 237.1 (1.4) & 1338.2 (13.2) \\ \hline
\end{tabular}
\caption{Mean and standard deviation of random and upper-bound performance (with $N=10, K=3$) across different subreddits.}
\label{table:subreddit-basic-performance}
\end{table}

\begin{table}
\small
\centering
\begin{tabular}{| l | c | c |} \hline
\bf K & \bf Random & \bf Upper bound \\ \hline
2 & 201.0 (2.1) & 1991.3 (2.9) \\ \hline
3 & 321.3 (7.0) & 2109.0 (16.5) \\ \hline
4 & 447.1 (10.8) & 2206.6 (8.2) \\ \hline
5 & 561.3 (18.8) & 2298.0 (29.1) \\ \hline
\end{tabular}
\caption{Mean and standard deviation of random and upper-bound performance on askscience, with $N=10$ and $K=2,3,4,5$.}
\label{table:askscience-basic-performance}
\end{table}

In all our experiments we set $N=10$. Explicitly representing all $N$-choose-$K$ actions requires a lot of memory and does not scale up. We therefore use a variant of Q-learning: when taking the max over possible next-actions, we instead randomly subsample $m'$ actions and take the max over them. We set $m'=10$ throughout our experiments. This heuristic technique works well in our experiments.
%%MOnew: How do we know it ``works well''? Can we instead say that it has been used in other studies and cite some?

For text preprocessing we remove punctuation and lowercase capital letters. For each state $s_t$ and comment $c_t^i$, we use a bag-of-words representation with the same vocabulary in all networks. The vocabulary contains the most frequent 5,000 words; the out-of-vocabulary rate is 7.1\%.

In terms of the Q-learning agent, fully-connected neural networks are used for text embeddings. The network has $L=2$ hidden layers, each with 20 nodes, and model parameters are initialized with small random numbers. $\epsilon$-greedy is used for exploration-exploitation, and we keep $\epsilon=0.1$ throughout online training and testing. We pick the discount factor $\gamma=0.9$. During online training, we use experience replay \cite{lin1992self} and the memory size is set to 10,000 tuples of $(s_t,a_t,r_{t+1},s_{t+1})$. For each experience replay, 500 episodes are generated and stored in a first-in-first-out fashion, and multiple epochs are trained for each model. Minibatch stochastic gradient descent is implemented with a batch size of 100. The learning rate is kept constant: $\eta_t=0.000001$.

The proposed methods are compared with three baseline models: Linear, per-action DQN (PA-DQN), and DRRN. For both Linear and PA-DQN, the state and comments are concatenated as an input. For the DRRN, the state and comments are sent through two separate deep neural networks. However, in our baselines, we do not explicitly model how values associated with each comment are combined to form the action value. For the DRRN baseline and proposed methods (DRRN-Sum and DRRN-BiLSTM), we use an inner product as the pairwise interaction function.

\subsection{Experimental Results}
\begin{figure}[t]
\centerline{
		\includegraphics[width=0.5\textwidth]{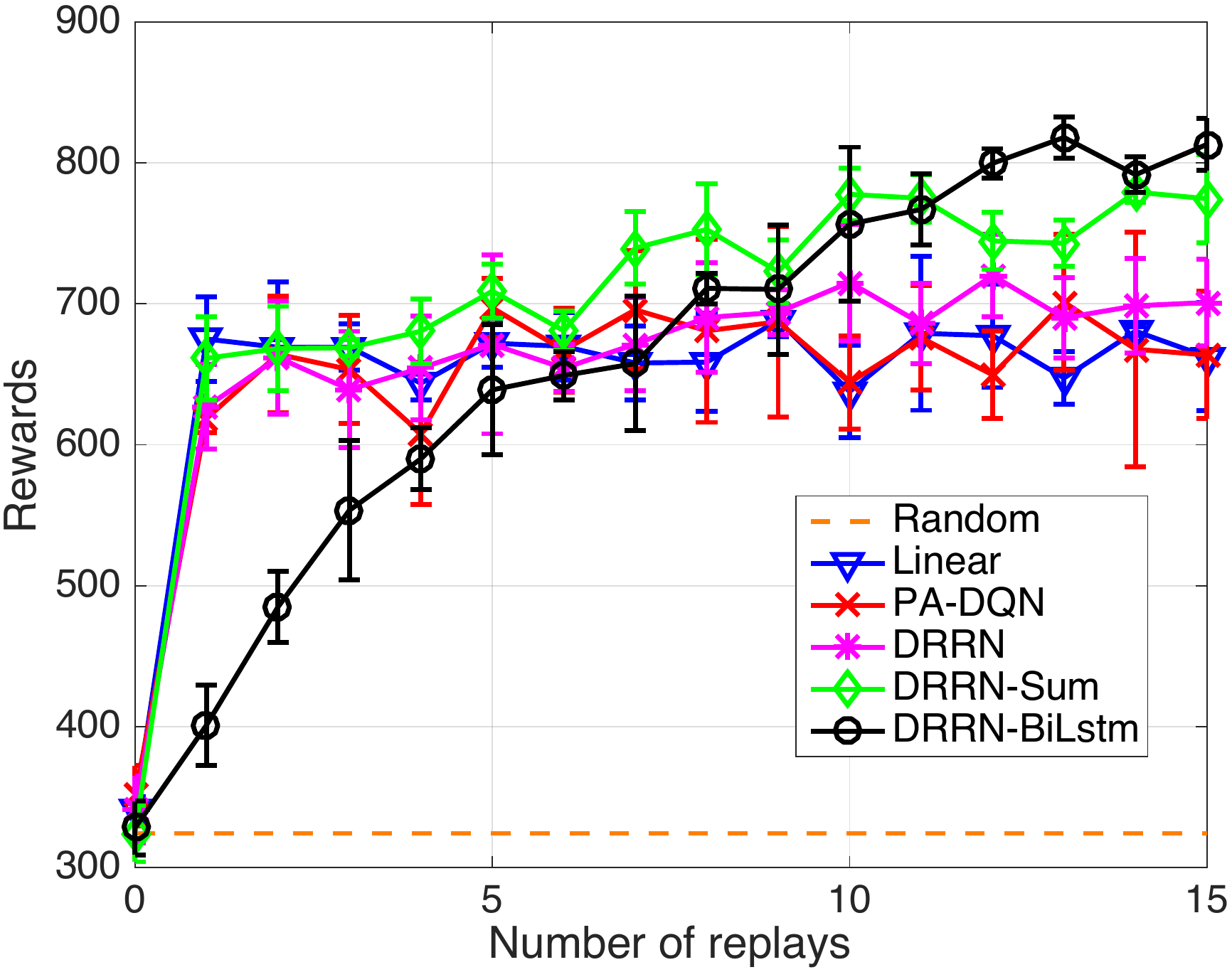}
 } 
  \caption{Learning curves of baselines and proposed methods on ``askscience''}
  \label{Fig:LearningCurve}
\end{figure}

In Figure \ref{Fig:LearningCurve} we provide learning curves of different models on the askscience subreddit during online learning. In this experiment, we set $N=10, K=3$. Each curve is obtained by averaging over 3 independent runs, and the error bars are also shown. All models start with random performance, and converge after approximately 15 experience replays. The DRRN-Sum converges as fast as baseline models, with better converged performance. DRRN-BiLSTM converges slower than other methods, but with the best converged performance.

After we train all the models on the training set, we fix the model parameters and apply (deploy) on the test set, where the models predict which action to take but no reward is shown until evaluation. The test performance is averaged over 1000 episodes, and we report mean and standard deviation over 5 independent runs.

\begin{table*}[t]
\small
\centering
\begin{tabular}{| l | c | c | c || c | c |} \hline
\bf K & \bf Linear & \bf PA-DQN & \bf DRRN & \bf DRRN-Sum & \bf DRRN-BiLSTM \\ \hline
2 & 553.3 (2.8) & 556.8 (14.5) & 553.0 (17.5) & 569.6 (18.4) & \bf 573.2 (12.9) \\ \hline
3 & 656.2 (22.5) & 668.3 (19.9) & 694.9 (15.5) & 704.3 (20.1) & \bf 711.1 (8.7) \\ \hline
4 & 812.5 (23.4) & 818.0 (29.9) & 828.2 (27.5) & 829.9 (13.2) & \bf 854.7 (16.0) \\ \hline
5 & 861.6 (28.3) & 884.3 (11.4) & 921.8 (10.7) & 942.3 (19.1) & \bf 980.9 (21.1) \\ \hline
\end{tabular}
\caption{On askscience, average karma scores and standard deviation of baselines and proposed methods (with $N=10$)}
\label{table:askscience-model-performance}
\end{table*}

On askscience, we try multiple settings with $N=10,\ K=2,3,4,5$ and the results are shown in Table \ref{table:askscience-model-performance}. Both DRRN-Sum and DRRN-BiLSTM consistently outperform baseline methods. The DRRN-BiLSTM performs better with larger $K$, probably due to the greater chance of redundancy in combining more sub-actions.

\begin{figure*}[t]
\centerline{
		\includegraphics[width=0.95\textwidth]{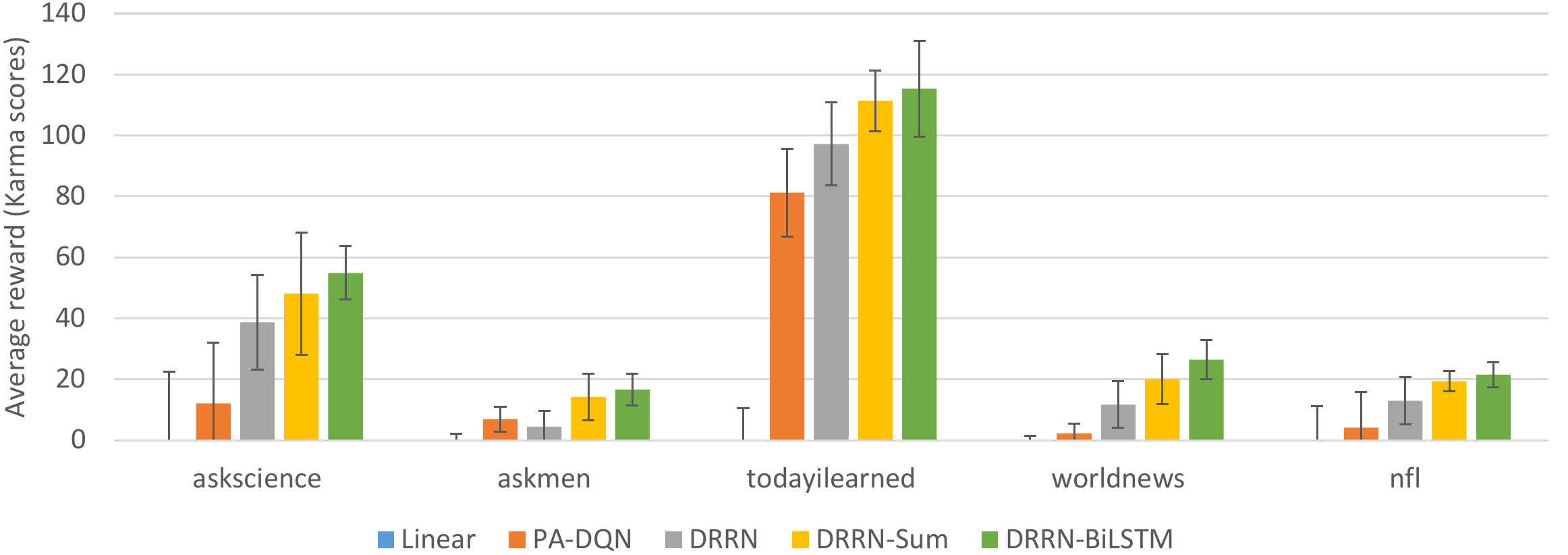}
 } 
  \caption{Average karma score gains over the linear baseline and standard deviation across different subreddits (with $N=10, K=3$).}
  \label{Fig:subreddit-model-performance}
\end{figure*}

We also perform online training and test across different subreddits. With $N=10, K=3$, the test performance gains over the linear baseline are shown in Figure \ref{Fig:subreddit-model-performance}. Again, the test performance is averaged over 1000 episodes, and we report mean and standard deviation over 5 independent runs. The findings are consistent with those for askscience. Since different subreddits may have very different karma scores distributions and language style, this suggests the algorithms apply to different text genres.

%\begin{table*}[t]
%\small
%\centering
%\begin{tabular}{| l | c | c | c || c | c | c |} \hline
%\bf Subreddit & \bf Linear & \bf PA-DQN & \bf DRRN & \bf DRRN-Sum & \bf DRRN-Pooling & \bf DRRN-BiLSTM \\ \hline
%askscience & 656.2 (22.5) & 668.3 (19.9) & 694.9 (15.5) & 704.3 (20.1) & 668.3 (19.5) & \bf 711.1 (8.7) \\ \hline
%askmen & 133.1 (2.1) & 139.9 (4.1) & 137.6 (5.1) & 147.3 (7.6) & 141.3 (6.1) & \bf 149.7 (5.2) \\ \hline
%todayilearned & 491.6 (10.6) & 572.8 (14.5) & 588.8 (13.6) & 602.9 (10.0) & 598.2 (12.6) & \bf 606.9 (15.8) \\ \hline
%worldnews & 235.5 (1.4) & 237.7 (3.3) & 247.2 (7.6) & 255.5 (8.2) & 249.5 (5.7) & \bf 262.0 (6.5) \\ \hline
%nfl & 327.8 (11.1) & 332.0 (11.7) & 340.8 (7.7) & 347.2 (3.4) & 338.8 (3.8) & \bf 349.3 (4.1) \\ \hline
%\end{tabular}
%\caption{Average Karma scores and standard deviation of baselines and proposed methods across different subreddits (with $N=10, K=3$).}
%\label{table:subreddit-model-performance}
%\end{table*}

\begin{table}
\small
\centering
\begin{tabular}{| c | c | c |} \hline
\bf K & \bf DRRN-Sum & \bf DRRN-BiLSTM \\ \hline
2 & 538.5 (18.9) & \bf 551.2 (10.5) \\ \hline
4 & 819.1 (14.7) & \bf 829.9 (11.1) \\ \hline
5 & 921.6 (15.6) & \bf 951.3 (15.7) \\ \hline
\end{tabular}
\caption{On askscience, average karma scores and standard deviation of proposed methods trained with $K=3$ and test with different $K$'s}
\label{table:askscience-proposed-generalization}
\end{table}

In actual model deployment, a possible scenario is that users may have different requests. For example, a user may ask the agent to provide $K=2$ discussion threads on one day, due to limited reading time, and ask the agent to provide $K=5$ discussion threads on the other day. For the baseline models (Linear, PA-DQN, DRRN), we will need to train separate models for different $K$'s. The proposed methods (DRRN-Sum and DRRN-BiLSTM), on the other hand, can easily handle a varying $K$. To test whether the performance indeed generalizes well, we train proposed models on askscience with $N=10, K=3$ and test them with $N=10, K\in {2, 4, 5}$, as shown in Table \ref{table:askscience-proposed-generalization}. Compared to the proposed models that are specifically trained for these $K$'s (Table \ref{table:askscience-model-performance}), the generalized test performance indeed degrades, as expected. However, in many cases, our proposed methods still outperform all three baselines (Linear, PA-DQN and DRRN) that are trained specifically for these $K$'s. This shows that the proposed methods can generalize to varying $K$'s even if it is trained on a particular value of $K$.

%2 & 553.3 (2.8) & \bf 556.8 (14.5) & 538.5 (18.9) & 542.5 (10.6) & 551.2 (10.5) \\ \hline
%4 & 812.5 (23.4) & 818.0 (29.9) & 819.1 (14.7) & 774.5 (14.3) & \bf 829.9 (11.1) \\ \hline
%5 & 861.6 (28.3) & 884.3 (11.4) & 921.6 (15.6) & 890.2 (16.7) & \bf 951.3 (15.7) \\ \hline

In Table \ref{table:q_value_examples}, we show an anecdotal example with state and sub-actions. The two sub-actions are strongly correlated and have redundant information. By combining the second sub-action compared to choosing just the first sub-action alone, DRRN-Sum and DRRN-BiLSTM predict 86\% and 26\% relative increase in action-value, respectively. Since these two sub-actions are highly redundant, we hypothesize DRRN-BiLSTM is better than DRRN-Sum at capturing interdependency between sub-actions.

%predicted Q-values from DRRN-Sum and DRRN-BiLSTM. The action column consists of combined comment/sub-action indices. We also compute a karma score by accumulating all comments in the subtrees associated with each action. Although the total Karma score in the subtrees has different range compared to Q-values (because Q-values are expected score of multiple threads, which are parts of subtrees), we use the relative increase as a guidance for evaluating predicted Q-values. For both subtree Karma score and predicted Q-values, we divide each value by its first action's, which takes only the first sub-action. In this example when combining the second sub-action (with redundant information as in the first sub-action), the DRRN-BiLSTM works relatively well, but DRRN-Sum's independence assumption predicts a larger increase.

\begin{table}[t]
\small
\centering
\begin{tabular}{| p{7.7cm} |} \hline
\bf State text (partially shown) \\ \hline
Are there any cosmological phenomena that we strongly suspect will occur, but the universe just isn't old enough for them to have happened yet? \\ \hline
\bf Comments (sub-actions) (partially shown) \\ \hline
[1] White dwarf stars will eventually stop emitting light and become black dwarfs. [2] Yes, there are quite a few, such as: White dwarfs will cool down to black dwarfs. \\ \hline
\end{tabular}
\caption{An example state and its sub-actions}
\label{table:q_value_examples}
\end{table}

%\begin{table*}[t]
%\centerline{
%		\includegraphics[width=1.0\textwidth]{q_value_examples_v3.pdf}
% } 
%  \caption{Predicted Q-value examples. Karma scores and Q-values are normalized by dividing the first action to show relative increase when combining more sub-actions.}
%\label{table:q_value_examples}
%\end{table*}

%\begin{table*}
%\small
%\centering
%\begin{tabular}{| p{3.0cm} | p{4.0cm} | p{1.1cm} || p{1.8cm} | p{1.75cm} | p{2.15cm} |} \hline
%state text (partial) & comment text (sub-action) (partial) & action & Subtree Karma score & DRRN-Sum predicted Q & DRRN-BiLSTM predicted Q \\ \hline
%\multirow{3}{*}{\parbox{3.0cm}{Are there any cosmological phenomena that we strongly suspect will occur, but the universe just isn't old enough for them to have happened yet?}} & \multirow{3}{*}{[1] White dwarf stars will } & $(1)$ & 1.0 & 1.0 & 1.0 \\ \hhline{~~----}
%& & $(1, 2)$ & 1.002 & 1.862 & 1.260 \\ \hhline{~~----}
%& & $(1, 2, 3)$ & 1.162 & 5.641 & 1.620 \\ \hline
%\end{tabular}
%\caption{On askscience, average Karma scores and standard deviation of DRRN-Sum and DRRN-BiLSTM methods trained with $K=3$ and test with different $K$, comparing with Linear and PA-DQN baseline methods. The baseline methods are specifically trained for different $K$s.}
%\label{table:q_value_examples}
%\end{table*}

\section{Conclusion}
\label{sec:concl}
In this paper we introduce a new reinforcement learning task associated with predicting and tracking popular threads on Reddit. The states and actions are all described by natural language so the task is useful for language studies. We then develop novel deep Q-learning architectures to better model the state-action value function with a combinatorial action space. The proposed DRRN-BiLSTM method not only performs better across different experimental configurations and domains, but it also generalizes well for scenarios where the user can request changes in the number tracked.

This work represents a first step towards addressing the popularity prediction and tracking problem. While performance of the system beats several baselines, it still falls far short of the oracle result. Prior work has shown that timing is an important factor in predicting popularity \cite{LampeRes04,Jaech2015}, and all the proposed models would benefit from incorporating this information. Another variant might consider short-term reactions to a comment, if any, in the update window. It would also be of interest to explore implementations of backtracking in the sub-action space (incurring a cost), in order to recommend comments that were not selected earlier but have become highly popular. Lastly, it will be important to study principled solutions for handling the computational complexity of the combinatorial action space.
%In this paper we propose a new large-scale reinforcement learning task by tracking popular threads on Reddit. The states and actions are all described by natural language so it will be useful for language studies. We also develop novel deep Q-learning architectures for better modeling the state-action value function in a combinatorial action scenario. The proposed methods can not only perform better across different experimental configurations and various domains, but also generalize well when user request changes. Future work includes: (i) adding non-language features such as timing to improve popularity tracking performance, and (ii) principled solutions for handling computational complexity in a large action space.
%
%Other future work to mention: backtracking in the sub-action space, leveraging immediate reactions in prediction

\bibliography{refs}
%\bibliography{emnlp2016}
\bibliographystyle{emnlp2016}

\end{document}